\documentclass[conference]{IEEEtran}
\IEEEoverridecommandlockouts
\usepackage{cite}
\usepackage{amsmath,amssymb,amsfonts}
\usepackage{algorithmic}
\usepackage{graphicx}
\usepackage{textcomp}
\usepackage{array}  %
\usepackage{xcolor}
\usepackage{authblk}
\usepackage{multirow}  

\def\BibTeX{{\rm B\kern-.05em{\sc i\kern-.025em b}\kern-.08em
    T\kern-.1667em\lower.7ex\hbox{E}\kern-.125emX}}
\begin{document}

\title{ Conformal Prediction on Quantifying Uncertainty of Dynamic Systems \\
}

\author[1,2,3]{Aoming Liang\thanks{Email: liangaoming@westlake.edu.cn}}
\author[2]{Qi Liu*\thanks{Email: liuqi76@westlake.edu.cn}}
\author[3]{Lei Xu\thanks{Email: lei.xu@tuni.fi}}
\author[4]{Fahad Sohrab\thanks{Email: fahad.sohrab@tuni.fi}}
\author[2]{Weicheng Cui\thanks{Email: cuiweicheng@westlake.edu.cn}}
\author[2]{Changhui Song\thanks{Email: songchanghui@westlake.edu.cn}}
\author[8]{Moncef Gabbouj*\thanks{Email: moncef.gabbouj@tuni.fi}}
\affil[1]{Zhejiang University}
\affil[2]{School of Engineering, Westlake Universty}
\affil[3]{Department of Computer Sciences, Tampere Universty}
\maketitle

\begin{abstract}
Numerous studies have focused on learning and understanding the dynamics of physical systems from video data, such as spatial intelligence.  Artificial intelligence requires quantitative assessments of the uncertainty of the model to ensure reliability. However, there is still a relative lack of systematic assessment of the uncertainties, particularly the uncertainties of the physical data. Our motivation is to introduce conformal prediction into the uncertainty assessment of dynamical systems, providing a method supported by theoretical guarantees. This paper uses the conformal prediction method to assess uncertainties with benchmark operator learning methods. We have also compared the Monte Carlo Dropout and Ensemble methods in the partial differential equations dataset, effectively evaluating uncertainty through straight roll-outs, making it ideal for time-series tasks.
\end{abstract}


\section{Introduction}
Video data \cite{bourdon2003pde,chen2007video,wu2023disentangling} can be represented by a set of time-dependent partial differential equations (PDEs), where the solutions correspond to the dynamic behavior. Similarly, image data can be interpreted as the solutions to time-independent partial or ordinary differential equations \cite{andris2016proof}. Exploring the uncertainties in PDEs using various methods is highly beneficial, particularly for tasks such as video understanding and generation \cite{wang2023internvid}.

PDEs serve as a fundamental framework for physical phenomena, with significant applications in weather forecasting \cite{fathi2022big}, nuclear fusion \cite{sadik2024back}, and molecular dynamics simulations \cite{zhang2023artificial}. Extensive research \cite{liang2024mixed,li2020fourier,kovachki2023neural} has been undertaken on how to learn the PDEs by artificial intelligence, especially those incorporating temporal evolution data, which can be conceptualized as sequences of video data structured on uniform or non-uniform grids. Given the deterministic nature of physical laws, the evolution processes should be deterministic.  The leading work primarily focuses on neural operator learning \cite{azizzadenesheli2024neural}.

However, the inherent uncertainties in artificial intelligence models \cite{abdar2021review} pose challenges in quantifying the uncertainties during the learning processes and exploring whether calibration can mitigate these uncertainties. Uncertainty quantification provides insights into the credibility or confidence level of predictive outcomes. The researcher can better evaluate the model's stability, accuracy, and reliability by accurately quantifying uncertainties.

Classic uncertainty studies \cite{mouli2024using,yang2021b,herrmann2020deep} only focus on one-step predictions of models, neglecting the cumulative uncertainties during temporal evolution in the forward or inverse problems. 

Rotational invariance \cite{hollard1982rotational,kalra2021towards} is a fundamental concept in the realm of physical systems, underpinning the understanding of symmetry in nature. Its significance is rooted in the principle that physical laws should remain unchanged under rotational transformations of the coordinate system. This invariance not only highlights the universality of physical laws but also plays a crucial role in the development of theoretical frameworks across various fields of physics.

The main contributions of this work are as follows:

\begin{itemize}
    \item In this study, a complete quantitative assessment of uncertainty quantification methods is performed, including Monte Carlo dropout, conformal prediction, and ensemble.
    \item The rotation symmetry in physical systems served as an opinion for evaluating uncertainty, utilizing calibration datasets to eliminate uncertainties is feasible and aligns with physical interpretation.
    \item Through conformal prediction, our work enhances the interpretability of models for time series tasks and provides theoretical upper and lower bounds as guarantees.
\end{itemize}

\section{Related Work on uncertainty evaluation}

\textbf{Bayesian methods}: Reference \cite{bernardo2009bayesian} firstly proposed introduced the Bayesian criterion, which assigns a prior distribution to the parameters of neural networks. With the training data, it then computes posterior distributions of these parameters to quantify predictive uncertainty. Reference \cite{blundell2015weight} proposed the variational Bayesian inference in the neural network. 
In practice, the prior is often set to Gaussian distributions to convert the classical Bayesian to an optimization problem, and computing the posterior requires multiple iterations, with time-consuming in high-dimensional data. 

\begin{figure*}[h]
    \centering
    \includegraphics[width=\textwidth]{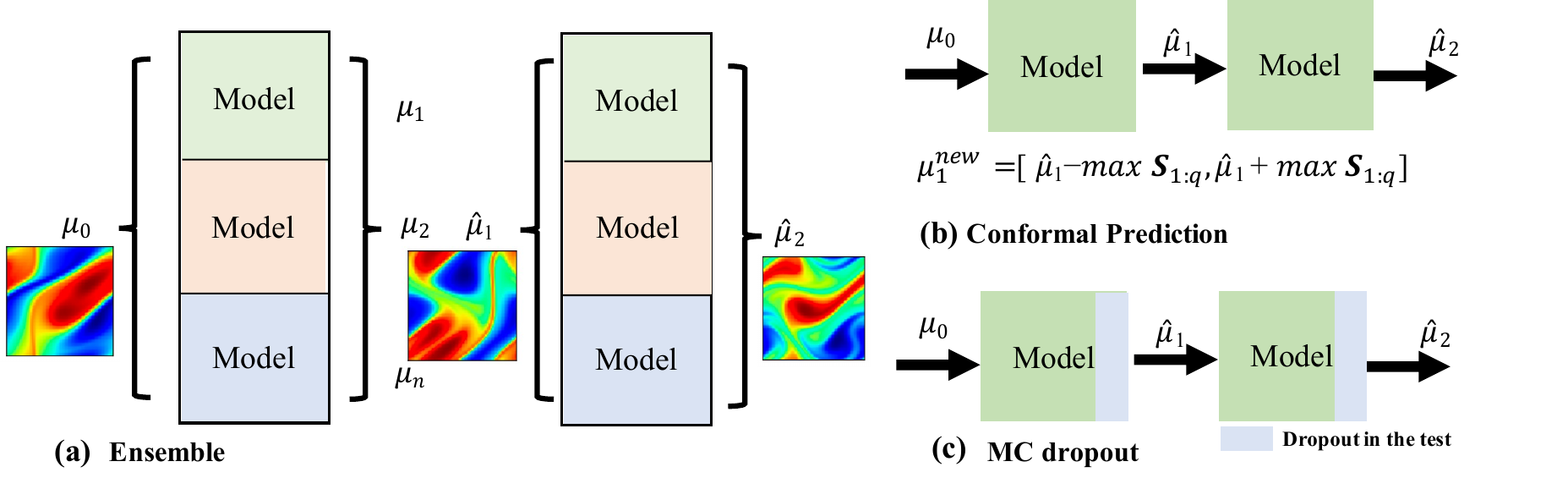}
    \caption{ The model is evaluated using three methods: (a) the ensemble method, (b) conformal prediction, and (c) Monte Carlo dropout. }
    \label{fig:roadmap}
\end{figure*}

\textbf{Monte Carlo dropout}: Based on classic Bayesian, reference \cite{gal2016dropout} suggested a Monte Carlo Dropout (MC Dropout) which acts as an approximation of Gaussian processes. With each forward pass, the network can randomly drop different nodes by utilizing a distinct random seed, resulting in diverse predictions for the same input, as depicted in Figure\ref{fig:roadmap}(c).

\textbf{Ensemble}:
Ensemble methods, as shown in Figure\ref{fig:roadmap}(a). Typically, models are assumed to be homoskedastic and exhibit the same variance distribution across all input data \cite{huang2017snapshot}. Researchers have recognized that performance improvements caused by assembling can lead to more stable predictions \cite{wenzel2020hyperparameter,jung2023simple}.

\textbf{Conformal prediction (CP)}: 
Conformal prediction \cite{sun2024conformal,angelopoulos2023conformal,straitouri2023improving}, as illustrated in Figure \ref{fig:roadmap}(b), is a relatively novel method for assessing uncertainty. Fundamentally, it achieves this by transforming an algorithm's predictions into prediction sets that guarantee robust coverage even for finite samples. Unlike traditional methods that might only assess the uncertainty at a single point, it leverages the information from the calibration dataset to ensure that the coverage probability of prediction intervals is not less than the chosen confidence level, regardless of the underlying distribution. This makes it particularly valuable in settings where accurate and reliable prediction intervals are crucial, such as in the issue of temporal dynamics in PDEs, where uncertainties can accumulate over time and symmetry.

\section{Methods}

In this section, we provide a conceptual introduction to the foundational theory of conformal prediction, outline the PDEs tasks, and explain how the conformal prediction can be applied to problems in this study.

\subsection{CP: Preliminary}

The core of CP is to construct a prediction set that provides uncertainty estimates for machine learning models. Let \((X_i, Y_i) \sim P\), \(i = 1, \dots, n\), represent \(n\) independent and identically distributed (i.i.d.) samples drawn from an unknown distribution \(P\), where:
\begin{itemize}
    \item \(X_i \in \mathcal{X}\) represents the feature/input space,
    \item \(Y_i \in \mathcal{Y}\) represents the response/output space,
    \item \(\mathcal{X} \subseteq \mathbb{R}^d\) and \(\mathcal{Y} \subseteq \mathbb{R}\), though this can be generalized to other domains.
\end{itemize}

Define a set of prediction scores:
\[
\hat{C}_n : \mathcal{X} \to \{\text{subsets of } \mathcal{Y}\},
\]
where \(\hat{C}_n(x)\) is the prediction set for a new input \(x \in \mathcal{X}\). The function \(\hat{C}_n\) is constructed such that it satisfies the following \textbf{finite-sample coverage property} for any nominal error level \(\alpha \in (0, 1)\):
\[
\mathbb{P}\left( Y_{n+1} \in \hat{C}_n(X_{n+1}) \mid (X_i, Y_i)_{i=1}^n \right) \geq 1 - \alpha,
\]
where:
\begin{itemize}
    \item \((X_{n+1}, Y_{n+1}) \sim P\) represents a new i.i.d. sample from the same distribution \(P\),
    \item the probability is computed over the randomness of the data \((X_i, Y_i)\), \(i = 1, \dots, n+1\).
\end{itemize}

This property ensures that the constructed prediction sets \(\hat{C}_n(x)\) achieve \textbf{valid coverage} with probability at least \(1 - \alpha\), even with finite training data.

\subsection{Description of PDEs task }

Consider a time-dependent PDE of the form:
\[
\frac{\partial u}{\partial t} = \mathcal{F}(u, \nabla u, \nabla^2 u, \dots),
\]
where \(u(t, x)\) is the state variable defined on a domain \(\Omega \subseteq \mathbb{R}^d\), and \(\mathcal{F}\) is a differential operator governing the dynamics of \(u\). The goal is to predict the state \(u_{n+1}\) at the next time step \(t_{n+1}\), given the state \(u_n\) at the current time step \(t_n\).

Let \((u_{n}, u_{n+1}) \sim P\), where:
\begin{itemize}
    \item \(u_n \in \mathcal{U}\) is the state of the system at time \(t_n\),
    \item \(u_{n+1} \in \mathcal{U}\) is the state at time \(t_{n+1}\),
    \item \(P\) represents the underlying distribution of states.
\end{itemize}

The prediction set \(\hat{C}_n(u_n)\) provides a range of possible states for \(u_{n+1}\), satisfying the finite-sample coverage property:
\[
\mathbb{P}\left( u_{n+1} \in \hat{C}_n(u_n) \mid (u_i, u_{i+1})_{i=1}^n \right) \geq 1 - \alpha,
\]
where \(\alpha\) is a predefined error tolerance level (e.g., \(\alpha = 0.05\) for 95\% confidence).

\subsection{CP in the PDEs task}

1. \textbf{Data Representation:} Collect training data \((u_i, u_{i+1})\), \(i = 1, \dots, n\), which consists of sequential states from a PDE simulation or experimental measurements.

2. \textbf{Nonconformity Measure:} Define a score function \(S(u_n, u_{n+1})\) that quantifies how unusual the next time state ${u}_{n+1}$ is relative to the previous state. For example:
\[
S(u_n, u_{n+1}) = \|u_{n+1} - \hat{u}_{n+1}\|_2,
\]
where \(\hat{u}_{n+1} = \mathcal{M}(u_n)\) is the predicted state using a learned model \(\mathcal{M}\) (e.g., neural operator or transformer model).

3. \textbf{Calibration:} Compute the nonconformity scores for the training data:
\[
S_i = S(u_i, u_{i+1}), \quad i = 1, \dots, n.
\]
Determine the \((1-\alpha)\)-quantile of these scores:
\[
Q_{1-\alpha} = \max(S_1, S_2, \dots, S_n).
\]

4. \textbf{Prediction Set:} For a new input state \(u_n\), construct the prediction set:
\[
\hat{C}_n(u_n) = \{v \in \mathcal{U} : S(u_n, v) \leq Q_{1-\alpha}\}.
\]
Then, \[
\mathbb{P}(u_{n+1} \in \hat{C}_n(u_n)) \geq 1 - \alpha
\]

\section{Expriments and results}

In this section, we will provide detailed results of our experiments, evaluation metrics, and the specific details of our experiments. To perform a thorough evaluation, we selected two of the most influential algorithms from each category, the Fourier neural operator (FNO) \cite{li2020fourier}, U-shaped neural operator (UNO) \cite{kovachki2023neural}, and tensorized Fourier neural operator (TFNO) \cite{kossaifi2023multi} for comparison. We quantitatively characterized all the evaluation results using the assessment metrics and isotonic regression developed by the Uncertainty Toolbox \cite{chung2021uncertainty}.

The core objective of this section is to address the following two questions. 

\begin{enumerate}
    \item \textbf{Uncertainty Quantification:} How do neural operators perform uncertainty quantification when applied to conformal prediction, ensemble methods, and MC dropout?
    
    \item \textbf{Symmetry Testing:} Given the inherent symmetries in physical systems, how do these models perform in symmetry-preserving tests?
    
\end{enumerate}
 
This study uses the fluid dataset in \cite{li2020fourier}, representing the video prediction task in the Navier Stokes (NS) equation.  In this paper, we have selected six models for our ensemble, utilizing the snapshot ensemble training method. For the dropout method, the selected rate was 0.05. All models discussed below have been pre-trained, with their parameters detailed in the Supplementary.  Given that the output of conformal prediction is the prediction interval \([a-b, a+b)\), where \(b\) is equal to \(Q_{1-\alpha}\), we have made a specific assumption about the standard deviation. Considering that physical phenomena in nature generally follow a Gaussian distribution, the standard deviation is assumed to be \(\frac{b}{z}\). Given that our chosen confidence interval is 95\%, the corresponding \(z\) value is 1.96.

\subsection{Forward and symmetry problems in turbulence dataset (NS)}

In the turbulence dataset, the forward problem and symmetry focus on evaluating the state of the fluid $u$ and $u'$ as it evolves. It contains $1200$ randomly sampled initial states and their trajectories, with $70\%$ used for training, $20\%$ for validation, and the remaining $10\%$ divided between test and calibration datasets.
\[
u_{1:10} \to u_{11:20}
\]

\[
u'_{1:10} \to u'_{11:20}
\]
\subsection{Results of Uncertainty of Neural Operators on the Forward Problems}

In this section, the illustrated and statistical results of different uncertainties are shown as follows.

\begin{figure}[h!]
    \centering
    \includegraphics[width=0.5\textwidth]{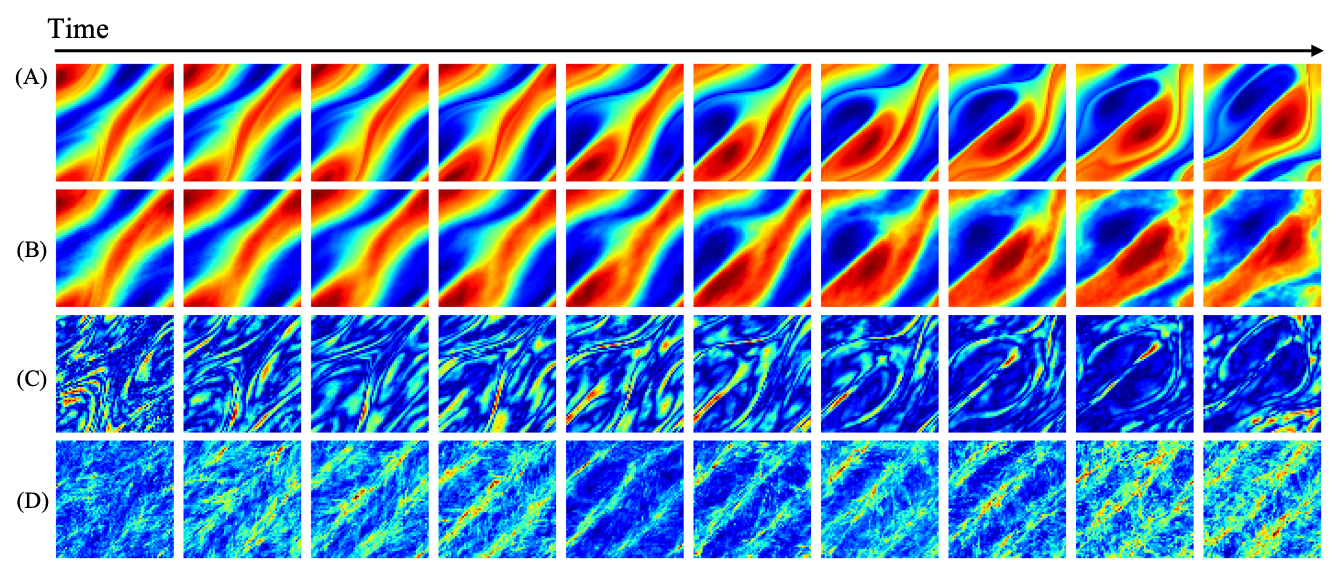} 
    \caption{Results of CP with $\alpha=0.05$ by FNO$_{128}$. Specifically, the first row corresponds to the ground truth, the second row is the model outputs, the third row is the absolute difference, while the fourth row, denoted as $\frac{Q_{1-\alpha}}{1.96}$, depicts the standard deviation across time. }
    \label{fig:cp} 
\end{figure}

\begin{figure}[h!] %
    \centering 
    \includegraphics[width=0.5\textwidth]{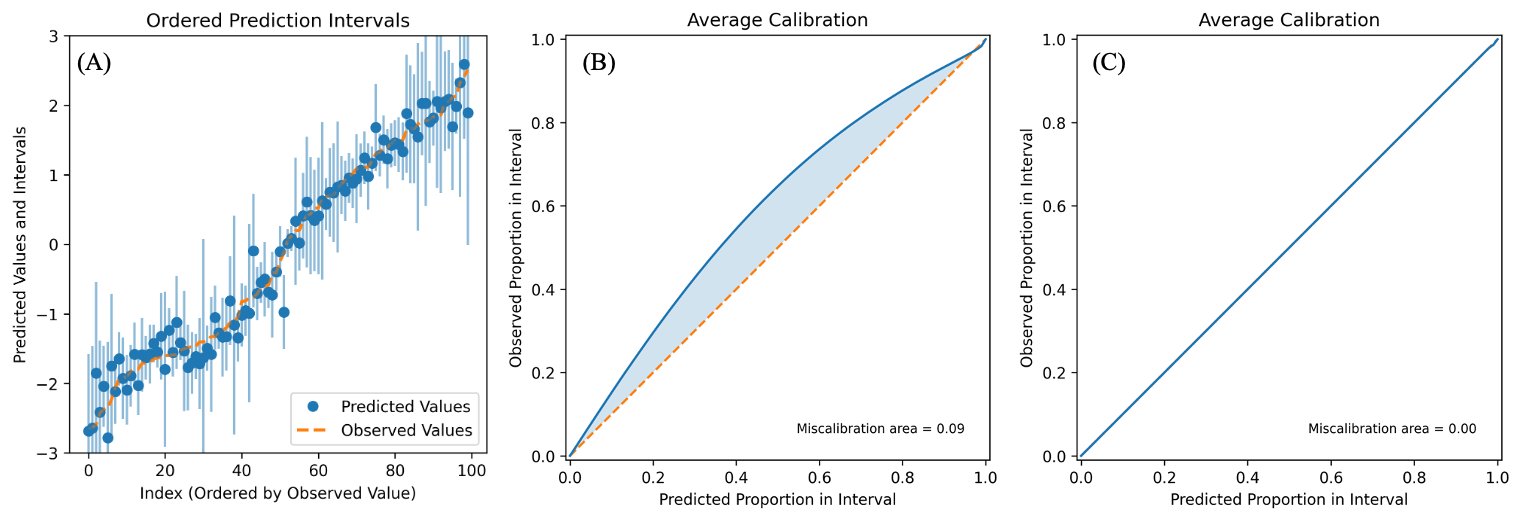} 
    \caption{The calibration results of CP. From left to right, the subfigure respectively represents the ordered prediction interval, the average calibration curve, and the re-calibrated curve after isotonic regression..  } 
    \label{fig:cp_cali} 
\end{figure}

\begin{figure}[h!]
    \centering
    \includegraphics[width=0.5\textwidth]{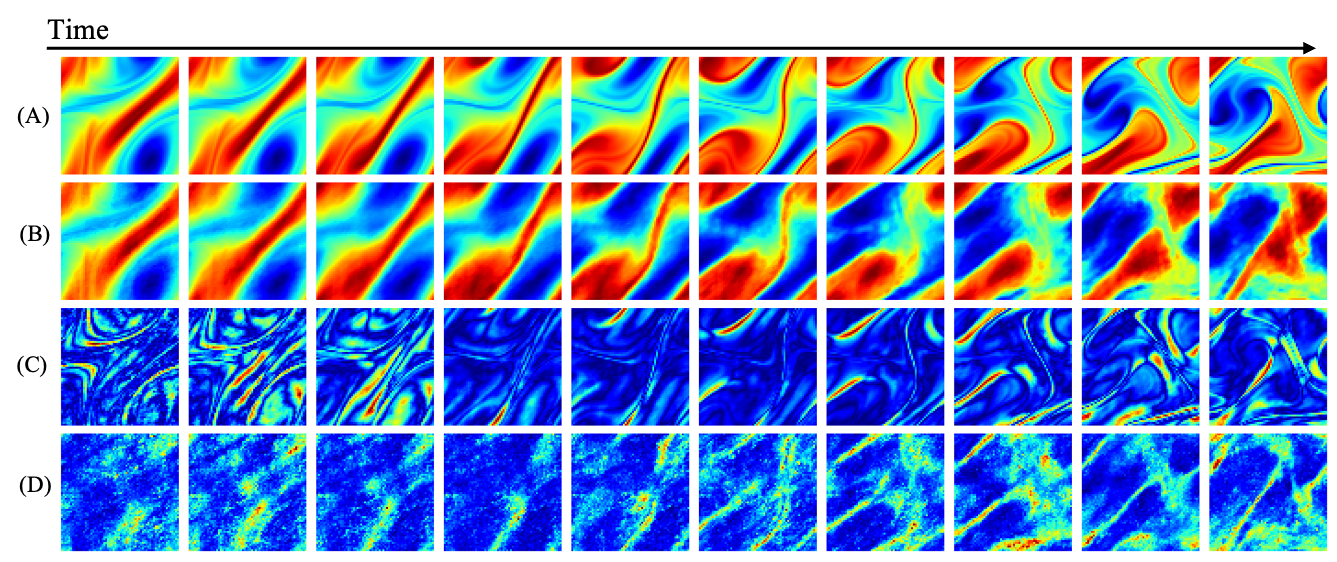} 
    \caption{Results of MC dropout by FNO$_{128}$. Specifically, the first row corresponds to the ground truth, the second row is the model outputs, the third row is the absolute difference, and the fourth row depicts the standard deviation across time.}
    \label{fig:5_mc} 
\end{figure}

\begin{figure}[h!] %
    \centering 
    \includegraphics[width=0.5\textwidth]{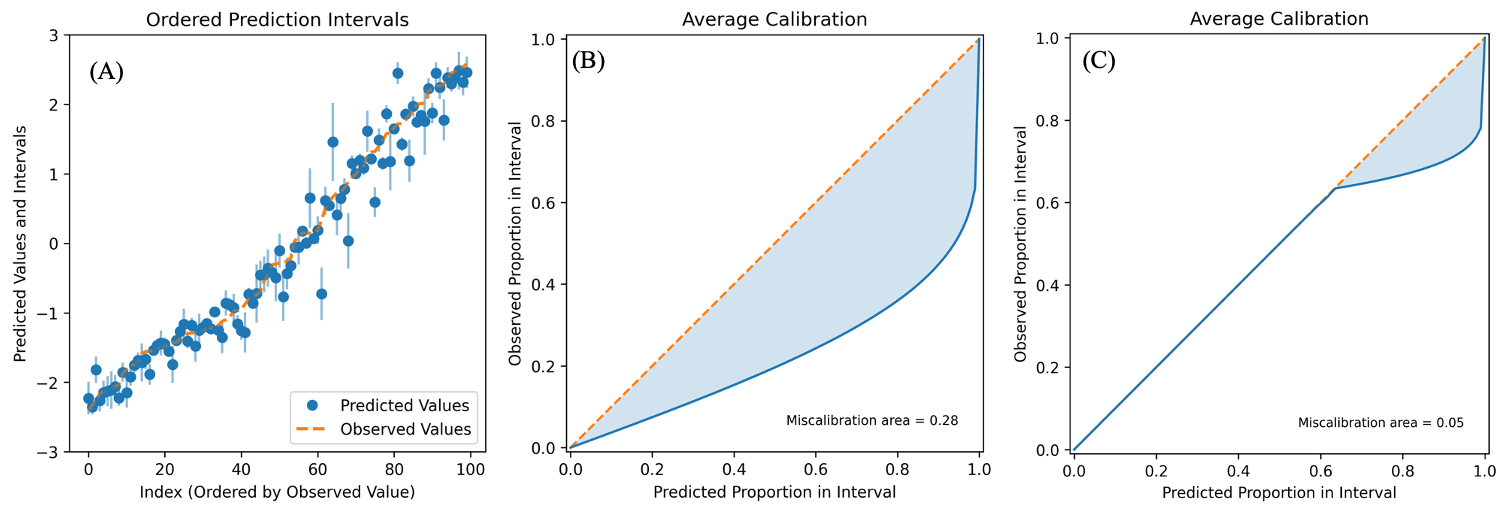} 
    \caption{The calibration results of MC dropout. From left to right, the subfigure respectively represents the ordered prediction interval, the average calibration curve, and the re-calibrated curve after isotonic regression. } 
    \label{fig:5_mc_cali} 
\end{figure}

\begin{figure}[h!]
    \centering
    \includegraphics[width=0.5\textwidth]{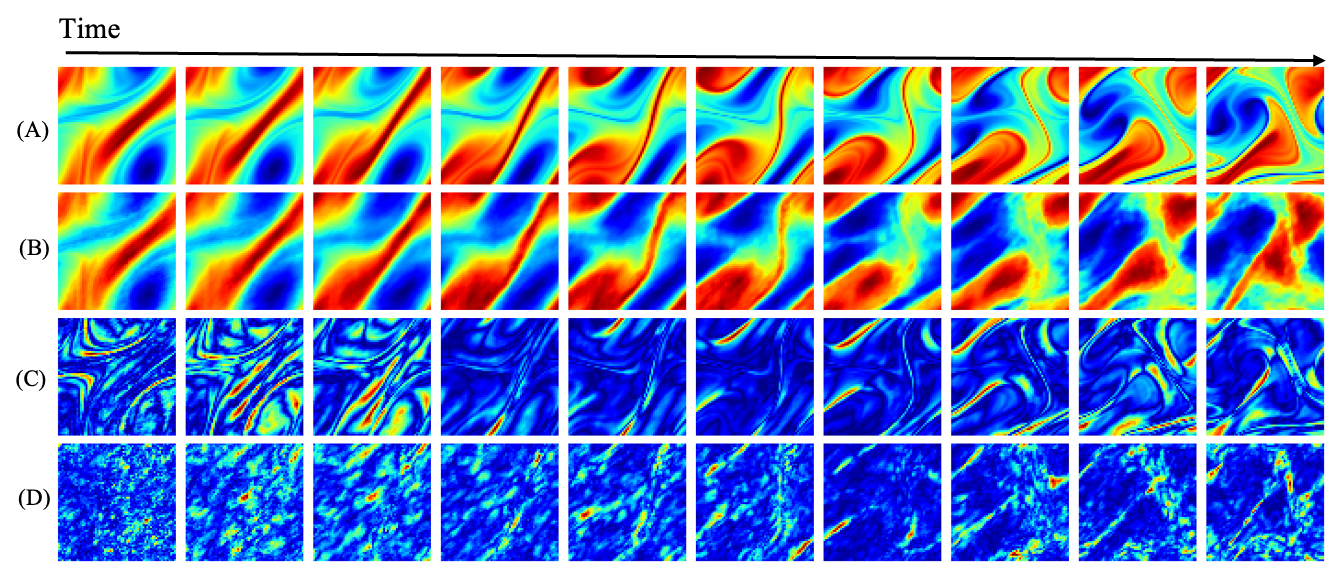} 
    \caption{Results of Ensemble by FNO$_{128}$. Specifically, the first row corresponds to the ground truth, the second row is the model outputs, the third row is the absolute difference, and the fourth row depicts the standard deviation across time.}
    \label{fig:6_ENS} 
\end{figure}

\begin{figure}[h!]
    \centering
    \includegraphics[width=0.5\textwidth]{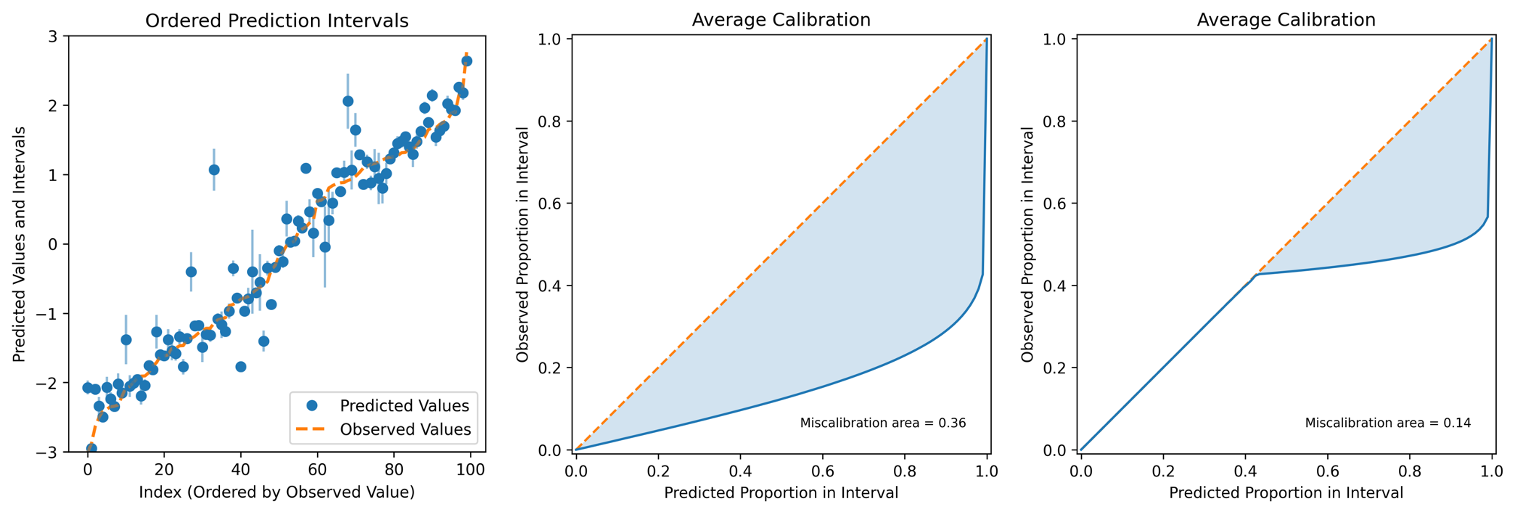} 
    \caption{The calibration results of Ensemble.  From left to right, the subfigure respectively represents the ordered prediction interval, the average calibration curve, and the re-calibrated curve after isotonic regression.}
    \label{fig:7_ENS_CALI} 
\end{figure}

\begin{table*}[h]
\centering
\renewcommand{\arraystretch}{1.1}  
\begin{tabular}{|>{\centering\arraybackslash}m{1.1cm}|>{\centering\arraybackslash}m{1.cm}|>{\centering\arraybackslash}m{1.1cm}|>{\centering\arraybackslash}m{1.1cm}|>{\centering\arraybackslash}m{1.1cm}|>{\centering\arraybackslash}m{2.3cm}|>{\centering\arraybackslash}m{2.3cm}|}
\hline
\textbf{Method} & \textbf{Operator} & \textbf{MAE} & \textbf{RMSE} & \textbf{Sharpness} & \textbf{Miscalibration Area (MA)} & \textbf{Recalibration Area (RA)} \\ \hline
\multirow{3}{*}{\centering Dropout} & FNO               &        0.24      &        0.38       &       0.11        &       0.26      &    0.05         \\ \cline{2-7} 
                                     & UNO               &     0.38         &    0.48           &    0.19           &       0.43      &    0.05         \\ \cline{2-7} 
                                     & TFNO              &  0.22            &     0.35          &    0.11           &     0.27        &     0.05        \\ \hline
\multirow{3}{*}{\centering Ensemble} & FNO               &       0.23       &      0.37         &       0.089        &      0.36       &    0.14         \\ \cline{2-7} 
                                     & UNO               &     0.35         &      0.54         &    0.22          &   0.27          &    0.08         \\ \cline{2-7} 
                                     & TFNO              &  0.21            &    0.34           &      0.11          &       0.33      &    0.11         \\ \hline
\multirow{3}{*}{\centering CP}       & FNO               &  0.21             &  0.33             &   0.41           &  0.093           &    0.0015     \\ \cline{2-7} 
                                     & UNO               &  0.37            &     0.56          &    0.22           &       0.065      &     0.00003        \\ \cline{2-7} 
                                     & TFNO              &    0.19          &    0.30           &      0.36         &      0.098      & 0.00005            \\ \hline
\end{tabular}
\caption{Different operators on the NS dataset uncertainty performance}
\label{tab1}
\end{table*}

Compared to the other two methods, as observed in Figure \ref{fig:7_ENS_CALI} and Figure \ref{fig:5_mc_cali}, the blue line of the CP method in Figure \ref{fig:cp_cali}(b) exceeds the orange diagonal line, indicating overconfidence in uncertainty assessment. This suggests that the variance predicted by the CP method is relatively loose, the essence cause may lie in the need to redesign the score function. However, this phenomenon can be mitigated by recalibration, as shown in Figure \ref{fig:cp_cali}(c), where the calibration error of CP is nearly zero post-recalibration, demonstrating a significant improvement in the uncertainty estimation by isotonic regression. From Table \ref{tab1}, it is evident that the CP method excels in calibration. This is because it uses its calibration data to ensure the accuracy of the prediction intervals. In contrast, methods like Dropout and Ensemble, although capable of providing information about model uncertainty, fail to effectively utilize calibration data, possibly leading to sub-par performance on test sets. Among the three uncertainty evaluation strategies, we compared the commonly used operator models horizontally. UNO demonstrated the poorest accuracy and uncertainty with the high sharpness distribution, while TFNO achieved the best results with the lowest MAE values. After calibration, the three models showed nearly identical performance in terms of miscalibration area for uncertainty. This result suggests that the CP method for confidence interval estimation outperforms both dropout and ensemble methods. CP is better suited for uncertainty estimation as it accounts for the physic law in the data, with \( Q_{1-\alpha} \) derived from the calibration set. The prediction intervals it generates are uniformly distributed and theoretically guaranteed.

\subsection{Results of Uncertainty of Neural Operators on the Symmetry testing}

Since dynamic systems, particularly PDEs, require consideration of rotational invariance, the predictions should correspondingly rotate when the input data is rotated. 

To evaluate this, we investigated whether the model could accurately predict the solution when presented with 90-degree rotational data. It is important to note that this analysis was primarily implemented within the CP method, as the calibration set also needed to be rotated accordingly.
\begin{table}[h]
\centering
\renewcommand{\arraystretch}{1.1}  
\begin{tabular}{|>{\centering\arraybackslash}m{1.5cm}|>{\centering\arraybackslash}m{1.cm}|>{\centering\arraybackslash}m{1.1cm}|>{\centering\arraybackslash}m{1.1cm}|>{\centering\arraybackslash}m{1.1cm}|>{\centering\arraybackslash}m{1.1cm}|}
\hline
\textbf{Operator} & \textbf{MAE} & \textbf{RMSE} & \textbf{Sharpness} & \textbf{MA} & \textbf{RA} \\ \hline
FNO & 0.21 & 0.33 & 0.91 & 0.31 & 0.017 \\ \cline{1-6}
UNO &0.37  &0.56 & 1.42 &0.27 & 0.004\\ \cline{1-6}
TFNO & 0.19& 0.30&0.92&0.32 &0.0007 \\ \hline
\end{tabular}
\caption{CP in symmetry Test}
\label{tab2}
\end{table}

\begin{figure}[h!]
    \centering
    \includegraphics[width=0.5\textwidth]{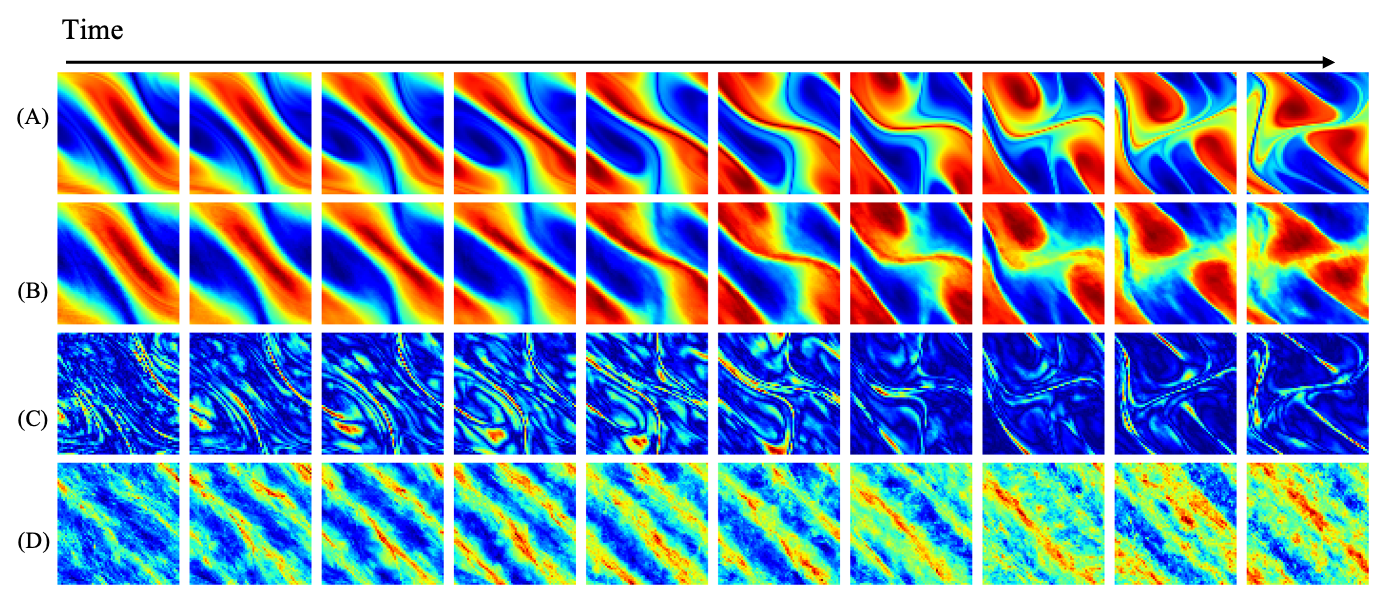} 
    \caption{Results of Symmetry test by FNO$_{128}$. Specifically, the first row corresponds to the ground truth, the second row is the model outputs, the third row is the absolute difference, and the fourth row depicts the standard deviation across time.}
    \label{fig:sy} 
\end{figure}

\begin{figure}[h!]
    \centering
    \includegraphics[width=0.5\textwidth]{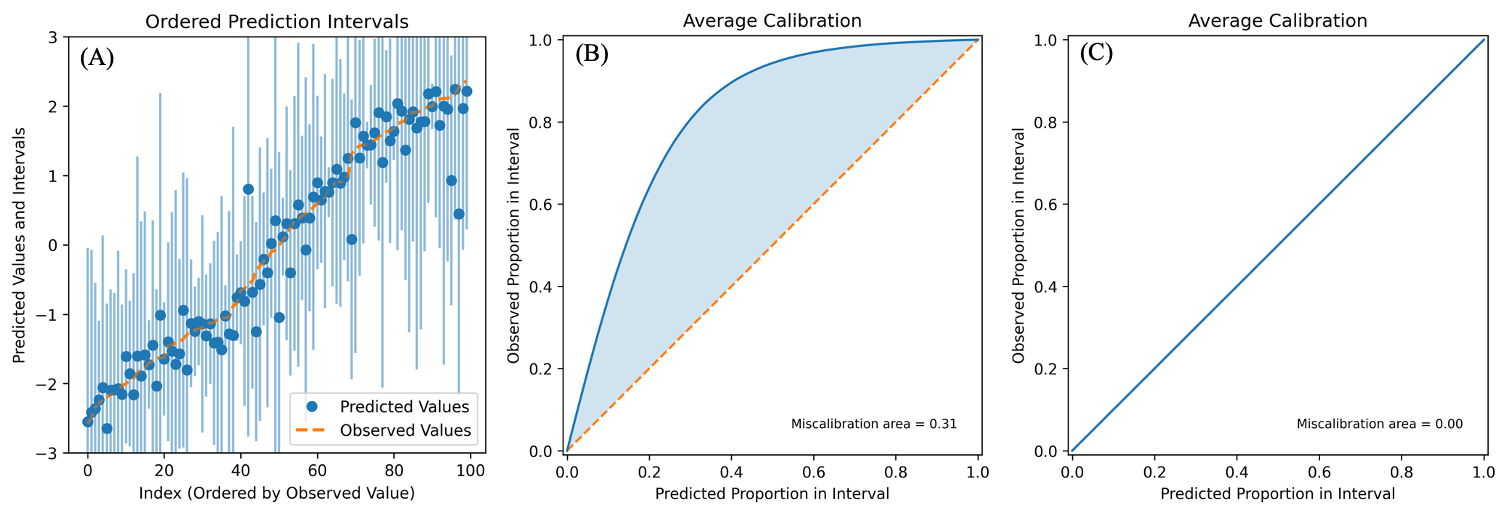} 
    \caption{Results of Symmetry test by FNO$_{128}$. From left to right, the subfigure respectively represents the ordered prediction interval, the average calibration curve, and the re-calibrated curve after isotonic regression.}
    \label{fig:sy_cli} 
\end{figure}

From Table \ref{tab1} and \ref{tab2}, it can be observed that after rotation, the statistical metrics MAE and RMSE remain nearly unchanged. The sharpness deviates from the original performance, highlighting an important characteristic of the representation of symmetry. It can be observed that UNO is not only incorrect, but also produces an overly concentrated predictive distribution. However, as shown in Figure \ref{fig:sy_cli}(a), the uncertainty of the predictions increases significantly compared to Figure \ref{fig:cp_cali}(a). 

This indicates that the uncertainty estimates of these models grow larger after rotational transformations. However, the output can still be recalibrated using isotonic regression, producing results that are largely consistent with the experiments in Section 3.2. In particular, the uncertainty estimation in Figure  \ref{fig:cp_cali}(c)  also rotates accordingly, demonstrating that the CP can correctly utilize the data from the calibration dataset for its calculations.

\section{Conclusion }

In this study, we employed conformal prediction to assess and understand model uncertainty in the complex partial differential equations tasks. We have derived uncertainty quantification curves for commonly used operator models through qualitative and quantitative comparative investigations. These can be broadly applied in fields such as scientific uncertainty and video evaluation, as well as in media forecast and interpretation. Future efforts will focus on how to reduce this uncertainty and control the training dynamic based on the uncertainty.

\bibliographystyle{IEEEbib}
\bibliography{ref}

\section{Appendix}

\subsection{Symmetry setting}
The experiments to investigate symmetry were conducted as follows: we trained the models using standard datasets, however, for the evaluation phase, the spatial positions of the input tensor data were systematically rotated by $90^\circ$ in the calibration and test dataset.

\subsection{Snapshot ensemble setting}

\subsubsection*{1. Cosine Annealing Learning Rate}

The learning rate \(\eta_t\) is adjusted using a cosine annealing schedule:

\[
\eta_t = \eta_{\text{min}} + \frac{1}{2} (\eta_{\text{max}} - \eta_{\text{min}}) \left( 1 + \cos\left(\frac{t}{T} \pi\right) \right)
\]

where:
\begin{itemize}
    \item \(\eta_{\text{max}}\) is the maximum learning rate at the start of the schedule; In this study \(\eta_{\text{max}}\) = 0.01
    \item \(\eta_{\text{min}}\) is the minimum learning rate at the end of the schedule; In this study \(\eta_{\text{min}}\) = 0.0001
    \item \(t\) is the current iteration step;  
    \item \(T\) is the total number of iterations in one cycle. In this study \(T\)  = 100
\end{itemize}

\subsubsection*{2. Snapshot Saving}

At the end of each learning rate cycle, when the learning rate reaches its minimum, the model is considered to have converged to a local optimum. The parameters of the model at this point are saved as a snapshot.

Let \(\theta_i\) denote the model parameters at the end of the \(i\)-th cycle. The set of all snapshot parameters is represented as:

\[
\Theta = \{\theta_1, \theta_2, \dots, \theta_M\}
\]

where \(M\) is the total number of snapshots saved during training.

\subsubsection*{3. Ensemble Prediction}

For inference, the predictions are made by averaging the outputs of all saved snapshots. The ensemble prediction is given by:

\[
\hat{y} = \frac{1}{M} \sum_{i=1}^M f_i(x)
\]

where \(f_i(x)\) is the prediction of the \(i\)-th snapshot model for input \(x\).

\begin{figure}[h]
    \centering
    \includegraphics[width=0.4\textwidth]{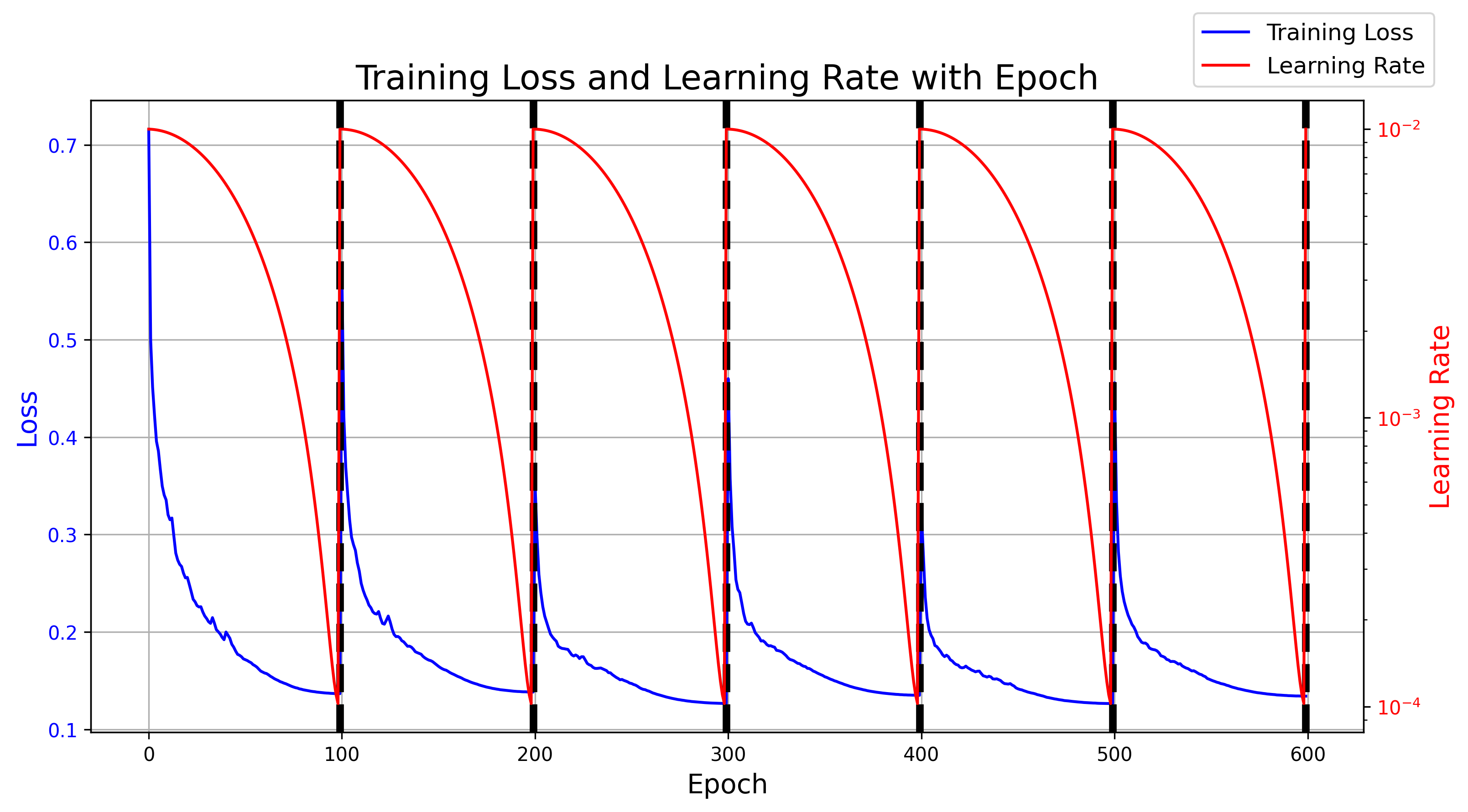}
    \caption{Training loss and learning rate across epochs about Ensemble method. The dashed lines indicate the times at which the models were saved, resulting in a total of six snap-shot models.}
    \label{fig:trainloss}
\end{figure}

\subsection{Baseline operators and hyper-parameters setting}
The baseline operator models selected in this study are as follows. The number of MC dropouts is set as $100$. The ensemble number is set to $M=6$.

\begin{table}[htbp]
    \centering
    \renewcommand{\arraystretch}{1.1}  
    \begin{tabular}{|c|c|c|}
        \hline
        Model & Property & Paras (M) \\ \hline
        \multirow{2}{*}{FNO} & Modes $(128, 128)$ & \multirow{2}{*}{17}  \\ \cline{2-2}
                             & Hidden Channels $16$ & \\ \hline
        \multirow{2}{*}{UNO} & Modes $(128, 128)$ & \multirow{2}{*}{6.8} \\ \cline{2-2}
                             & Scaling Layer $5$ & \\ \hline
        \multirow{2}{*}{TFNO} & Modes $(128, 128)$ & \multirow{2}{*}{7.9} \\ \cline{2-2}
                             & Rank $0.5$ & \\ \hline
    \end{tabular}
    \caption{Baseline Methods Setting}
    \label{tab:baseline_methods}
\end{table}

\subsection{Metric on the Uncertainty}

\paragraph{Sharpness} Sharpness is a measure of how narrow, concentrated, or peaked the predictive distribution is. Sharpness is evaluated solely based on the predictive distribution, and neither the datapoint nor the ground truth distribution is evaluated when measuring sharpness.

\paragraph{Miscalibration area} The Miscalibration area is a metric used to assess probability calibration, measuring the disparity between the probability distribution predicted by a model and the actual observed distribution.

\paragraph{MAE (Mean Absolute Error)} MAE typically refers to Mean Absolute Error. It is a common metric used to assess the quality of uncertainty estimation in predictive mea.

\paragraph{RMSE (Root Mean square Error)} RMSE is a widely used metric to quantify the accuracy of a model. It measures the square root of the average squared differences between the predicted and actual values. 
\end{document}